\documentclass[runningheads]{llncs}
\usepackage[T1]{fontenc}
\usepackage{amsmath}
\usepackage{graphicx}
\usepackage{tikz}
\usetikzlibrary{positioning, shapes.geometric, arrows.meta, fit, backgrounds, calc}
\usepackage{microtype}
\usepackage{xcolor}
\usepackage{hyperref}
\usepackage{url}
\usepackage[bottom]{footmisc}

\begin{document}
\title{MuTSE: A Human-in-the-Loop Multi-use Text Simplification Evaluator}

\author{Rareș-Alexandru Roșcan\inst{1} and  Gabriel Petre \inst{1} and Adrian-Marius Dumitran\inst{1,3,4}\orcidID{0009-0005-3547-5772} and Angela-Liliana Dumitran \inst{2}\orcidID{0009-0003-3590-9441}} 

\authorrunning{Rareș-Alexandru Roșcan}

% Corrected Institute/Email Section
\institute{University of Bucharest, Faculty of Mathematics and Computer Science, \\ Academiei 14, 010014, Bucharest, Romania\\
\email{marius.dumitran@unibuc.ro, radu.dita@gmail.com} \\ \and
Universitatea Creștină "Dimitrie Cantemir" \\
\email{angela.dumitran@gmail.com}
\and
Cu Drag si Sport SRL, Bucharest, Romania \and Softbinator Technologies, Bucharest, Romania}
\maketitle

\begin{abstract}
As Large Language Models (LLMs) become increasingly prevalent in text 
simplification, systematically evaluating their outputs across diverse 
prompting strategies and architectures remains a critical methodological 
challenge in both NLP research and Intelligent Tutoring Systems (ITS). 
Developing robust prompts is often hindered by the absence of structured, 
visual frameworks for comparative text analysis. While researchers typically 
rely on static computational scripts, educators are constrained to standard 
conversational interfaces --- neither paradigm supports systematic 
multi-dimensional evaluation of prompt-model permutations. To address 
these limitations, we introduce \textbf{MuTSE}\footnote{The project code and the demo have been made available for peer review at the following anonymized URL. \url{https://osf.io/njs43/overview?view_only=4b4655789f484110a942ebb7788cdf2a}}, an interactive 
human-in-the-loop web application designed to streamline the evaluation 
of LLM-generated text simplifications across arbitrary CEFR proficiency 
targets. The system supports concurrent execution of $P \times M$ 
prompt-model permutations, generating a comprehensive comparison matrix 
in real-time. By integrating a novel tiered semantic alignment engine 
augmented with a linearity bias heuristic ($\lambda$), MuTSE visually maps 
source sentences to their simplified counterparts, reducing the cognitive 
load associated with qualitative analysis and enabling reproducible, 
structured annotation for downstream NLP dataset construction.

\end{abstract}

\section{Introduction}

The adaptation of complex texts into accessible reading materials—known as text simplification—is a core component of Intelligent Tutoring Systems (ITS) and language learning applications. The rapid advancement of Large Language Models (LLMs) has provided educators and researchers with robust generative frameworks capable of tailoring texts to specific educational levels (e.g., CEFR A2, B1). Nevertheless, optimizing the interplay between LLM architectures and prompting strategies remains a significant methodological challenge. 

Currently, evaluating LLM outputs for text simplification is a fragmented and labor-intensive undertaking. Researchers and applied linguists frequently depend on standalone scripts or computational notebooks for text generation, approaches that lack scalability when systematically comparing multiple models against diverse prompts. This configuration yields a high-dimensional evaluation matrix of $P\times M$ (where $P$ denotes the number of prompts and $M$ the number of models). Furthermore, although existing methodologies employ conventional automated metrics (e.g., BLEU, SARI) for output evaluation, they typically lack the interactive, human-in-the-loop interfaces required for nuanced qualitative assessment. Consequently, tracking how a specific complex sentence is structurally altered across numerous generated variants imposes a substantial cognitive load on researchers.

To address these methodological constraints, we introduce MuTSE, an interactive, human-in-the-loop framework developed for the systematic evaluation of LLM-generated text simplifications. The system mitigates the operational complexity associated with model inference and prompt orchestration, enabling users to conduct parallel generation tasks across arbitrary permutations of local or cloud-based LLMs.

The primary contributions of this paper are:
\begin{itemize}
    \item \textbf{Parallel Comparative Workflow:} A system architecture capable of executing multi-dimensional ($P\times M$) text simplification tasks concurrently, presenting the results in a unified, side-by-side visualization.
    \item \textbf{Interactive Semantic Alignment:} A novel, tiered alignment engine featuring a linearity bias heuristic ($\lambda$) that visually maps original sentences to their simplified counterparts across all selected models in real-time.
    \item \textbf{Educational Annotation Framework:} An integrated suite of readability metrics (e.g., Flesch-Kincaid) and a customizable manual annotation system, enabling educators and linguists to score texts based on personalized criteria and export structured data for further analysis.
\end{itemize}

By bridging the gap between raw LLM capabilities and practical qualitative assessment, MuTSE provides an accessible environment for educators to select optimal texts and for NLP researchers to build high-quality annotated datasets.

\section{Related Work}

\subsection{Text Simplification Tools and Evaluation Interfaces}
While automated metrics drive large-scale benchmarking, the qualitative analysis of simplified text requires specialized software interfaces. At the programmatic level, toolkits like EASSE \cite{alva-manchego-etal-2019-easse} have successfully standardized the extraction of text simplification metrics (e.g., SARI, compression ratio) and the generation of automated evaluation reports. However, such packages remain strictly developer-centric, operating via command-line or code interfaces without visual alignment capabilities. 

On the visual interface front, contributions such as TS-ANNO \cite{stodden-kallmeyer-2022-ts} have introduced dedicated human-in-the-loop web environments for manually aligning, rating, and annotating text simplifications. While highly effective for post-hoc corpus creation, tools like TS-ANNO are primarily designed for manual annotation rather than live, concurrent multi-model generation. Furthermore, a recent comprehensive review of automatic text simplification tools \cite{espinosa-zaragoza-etal-2023-review} highlights a significant gap in the current ecosystem: a pervasive lack of publicly accessible resources that offer high customization and empower users to control the simplification process without technical barriers. 

MuTSE directly addresses these limitations. By combining the concurrent generation capabilities of modern LLMs with the automated metric extraction of programmatic toolkits and the visual semantic mapping of annotation platforms, MuTSE provides a unified, highly customizable, and accessible environment for rigorous text simplification research.

Within Intelligent Tutoring Systems, the problem of matching text complexity to learner proficiency has a long history. Systems such as REAP~\cite{brown2004reap} selected reading materials calibrated to individual vocabulary gaps, while Project LISTEN~\cite{mostow2002listen} demonstrated the pedagogical value of 
automated text-level assessment in supporting struggling readers. MuTSE extends this tradition by enabling educators to actively generate and comparatively evaluate simplified texts across proficiency targets such as CEFR A2 and B1, rather than merely selecting from pre-existing corpora.

\subsection{LLMs and the Evolution of Simplification Benchmarks}
The paradigm of automated text simplification has shifted dramatically toward the deployment of Large Language Models (LLMs). Recent studies \cite{feng2023sentencesimplificationlargelanguage} demonstrate that LLMs possess strong zero-shot capabilities for adapting complex texts. To systematically evaluate these capabilities, comprehensive benchmarks such as BLESS \cite{kew-etal-2023-bless} have been introduced, assessing dozens of off-the-shelf LLMs across multiple domains. While such benchmarks confirm that LLMs perform on par with state-of-the-art baselines, they also highlight the necessity of manual, qualitative analysis to truly gauge the diversity of edit operations and the educational quality of the generated text.

To evaluate these generative models computationally, researchers often rely on sophisticated, embedding-based automated metrics like BERTScore \cite{zhang2020bertscoreevaluatingtextgeneration}, which compute semantic similarity using contextual embeddings rather than exact lexical overlap. However, while metrics like BERTScore are excellent for static benchmarking, they aggregate performance into single numerical values. They lack the granular interpretability required by educators to visually map and validate multi-reference transformations. MuTSE bridges this methodological gap; it complements large-scale benchmarks by providing a real-time, human-in-the-loop environment where researchers can qualitatively analyze LLM outputs through visual semantic alignment, moving beyond opaque aggregate scores.

\section{System Architecture and Methodology}

To facilitate the near real-time evaluation of text simplification, MuTSE is built upon a decoupled, asynchronous client-server architecture. The backend is powered by Python and FastAPI, providing a high-performance REST API that orchestrates parallel LLM generation tasks via Together AI's serverless endpoints. On the frontend, a Vue.js 3 interface leverages reactive state management to handle multi-dimensional comparisons and client-side alignments. 

This infrastructure empowers researchers to systematically benchmark multiple LLMs and prompting strategies via an interactive, human-in-the-loop visualization framework. By automatically identifying and highlighting semantically corresponding sentences—utilizing a core NLP pipeline based on \textit{spaCy}, \textit{sentence-transformers}, and \textit{scikit-learn}—the platform alleviates the challenges of tracking complex transformations. Furthermore, a lightweight, local JSON-based persistence layer ensures maximum portability for educators without requiring complex database configurations. The following subsections detail the concurrency framework designed to compute $P \times M$ evaluation matrices efficiently, along with the alignment engine driving the comparative interface.

\subsection{Asynchronous Concurrency Model for Multi-Model Inference}

A fundamental constraint in multi-model evaluation is the latency overhead inherent to sequential network requests. To ensure the responsiveness required for an uninterrupted human-in-the-loop workflow, the architecture implements a highly concurrent execution paradigm. 

Upon request initiation, the orchestration layer provisions independent, distributed tasks for each prompt and model permutation. These processes are evaluated asynchronously, effectively restraining the overall computational latency to approximately $O(\max(t_i))$, where $t_i$ denotes the inference time of the slowest individual model iteration. However, an inherent characteristic of this concurrent approach is that the system's absolute responsiveness is strictly bounded by the highest latency within the queried batch. For instance, evaluating a highly optimized, low-latency model (e.g., Llama 3.3 Turbo) alongside a computationally heavy "reasoning" model will inevitably throttle the final matrix resolution time to match the latter's prolonged inference speed.

To maintain concurrency safety, the architecture strictly isolates state during the parallel execution phase. Each task independently manages text generation, semantic alignment computations, and metric extraction. Computationally intensive procedures, such as similarity matrix derivations, are strategically delegated to prevent blocking primary execution threads. The subsequent aggregation of the generated corpora and metadata is performed strictly after all concurrent tasks have reached completion, ensuring structural integrity without distributed state conflicts.

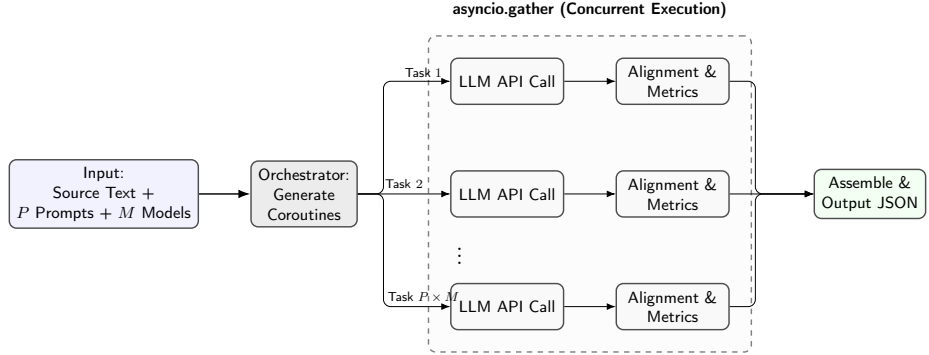
\begin{figure}[htbp]
    \centering
    % Scalăm doar dacă e absolut necesar, deși acum ar trebui să încapă nativ
    \resizebox{\textwidth}{!}{%
    \begin{tikzpicture}[>=Latex, font=\sffamily\small, node distance=0.8cm and 1cm]
    
        \tikzset{
            base/.style={draw=black!70, rectangle, rounded corners, align=center, minimum height=0.9cm, thick, inner sep=4pt},
            input/.style={base, fill=blue!5},
            process/.style={base, fill=gray!5},
            orchestrator/.style={base, fill=gray!15},
            lane/.style={process, minimum width=2.2cm},
            output/.style={base, fill=green!5},
            box/.style={draw=black!50, dashed, inner sep=12pt, rounded corners, fill=gray!2, thick},
            lane label/.style={font=\sffamily\scriptsize, above, midway}
        }
    
        % Nodes
        \node[input] (in) {Input:\\Source Text +\\$P$ Prompts + $M$ Models};
        \node[orchestrator, right=of in] (orch) {Orchestrator:\\Generate\\Coroutines};
        
        % Parallel Lanes (Combined Alignment & Metrics)
        \node[lane, right=1.8cm of orch, yshift=2.2cm] (api1) {LLM API Call};
        \node[lane, right=of api1] (align1) {Alignment \&\\Metrics};
        
        \node[lane, right=1.8cm of orch] (api2) {LLM API Call};
        \node[lane, right=of api2] (align2) {Alignment \&\\Metrics};
        
        \node[font=\Large, right=1.8cm of orch, yshift=-1.1cm] (dots) {$\vdots$};
        
        \node[lane, right=1.8cm of orch, yshift=-2.2cm] (api3) {LLM API Call};
        \node[lane, right=of api3] (align3) {Alignment \&\\Metrics};
        
        % Background Box for concurrent execution
        \begin{scope}[on background layer]
            \node[box, fit=(api1) (align1) (api3) (align3) (dots)] (gather) {};
            \node[above=6pt of gather.north, font=\sffamily\bfseries] (gatherLabel) {asyncio.gather (Concurrent Execution)};
        \end{scope}
        
        % Assemble and Output combined
        \node[output, right=1.2cm of gather] (out) {Assemble \&\\Output JSON};
        
        % Edges
        \draw[->, thick] (in) -- (orch);
        
        % Edges into the box
        \draw[->, rounded corners] (orch.east) -- ++(0.5,0) |- (api1.west) node[pos=0.8, above, scale=0.8] {Task $1$};
        \draw[->] (orch.east) -- (api2.west) node[lane label] {Task $2$};
        \draw[->, rounded corners] (orch.east) -- ++(0.5,0) |- (api3.west) node[pos=0.8, above, scale=0.8] {Task $P \times M$};
        
        % Edges inside lanes
        \draw[->] (api1) -- (align1);
        \draw[->] (api2) -- (align2);
        \draw[->] (api3) -- (align3);
        
        % Edges out of the box
        \draw[->, rounded corners] (align1.east) -- ++(0.5,0) |- (out.west);
        \draw[->] (align2.east) -- (out.west);
        \draw[->, rounded corners] (align3.east) -- ++(0.5,0) |- (out.west);
        
    \end{tikzpicture}%
    }
    \caption{Asynchronous parallel orchestration ($P \times M$) pattern demonstrating the concurrent execution of prompts across multiple LLMs.}
    \label{fig:orchestration}
\end{figure}

\subsection{Semantic Alignment Engine}

A central contribution of the system is the Semantic Alignment Engine, designed to visually map original sentences to their simplified counterparts. While text simplification predominantly exhibits a monotonic structural progression \cite{alva-manchego-etal-2020-asset,xu-etal-2016-optimizing}, standard cosine similarity measures applied to sentence embeddings frequently yield false positive alignments—often associating a simplified construct with a semantically congruent but positionally distant original sentence. To mitigate this alignment degradation, we propose a hierarchical fallback strategy augmented by a positional penalty heuristic.

\subsubsection{Hierarchical Embedding Strategy}
To guarantee operational resilience and hardware agnosticism, sentence embeddings are derived through a multi-level sequential cascade:
\begin{enumerate}
    \item \textbf{Primary Semantic Level:} Utilizes a condensed, 384-dimensional multilingual transformer-based model (\texttt{paraphrase-multilingual-MiniLM-L12-v2}) \cite{reimers2019sentencebertsentenceembeddingsusing} to compute the foundational cosine similarity matrix.
    \item \textbf{Secondary Lexical Level:} In scenarios where semantic embedding fails or yields null vectors, the algorithm reverts to a hybrid TF-IDF representation \cite{SALTON1988513}. This incorporates both word-level and character-level n-grams to capture foundational cross-lingual morphological similarities.
    \item \textbf{Tertiary Positional Level:} A definitive structural fallback that bypasses lexical similarity entirely, establishing alignments based strictly on normalized sequence positioning within the text documents.
\end{enumerate}

\subsubsection{Linearity Bias Assignment}
To align sentences accurately while avoiding the false positives inherent in purely semantic matching, we introduce a linearity factor ($\lambda$). The alignment score between an original sentence $i$ and a simplified sentence $j$ is calculated as:

\begin{equation}
\text{Score}(i, j) = \text{CosineSim}(S_{orig}^{(i)},S_{simp}^{(j)}) - \left| Pos_{rel}^{(i)} - Pos_{rel}^{(j)} \right| \times \lambda
\end{equation}

Where:
\begin{itemize}
    \item $S_{orig}^{(i)}$ and $S_{simp}^{(j)}$ represent the original and simplified sentences, respectively.
    \item $Pos_{rel}^{(i)} = \frac{i}{\max(N_{orig} - 1, 1)}$ and $Pos_{rel}^{(j)} = \frac{j}{\max(N_{simp} - 1, 1)}$ represent the normalized relative positions within their respective texts.
    \item $\lambda$ is the linearity factor, adjustable in real-time by the user between 0 and 2, with a default of 0.5.
\end{itemize}

The algorithm uses a many-to-one assignment, mapping each simplified sentence to the original sentence that yields the highest $\text{Score}(i, j)$. 

\begin{figure}[htbp]
    \centering
    % Prima imagine (lambda = 0)
    \includegraphics[width=\textwidth]{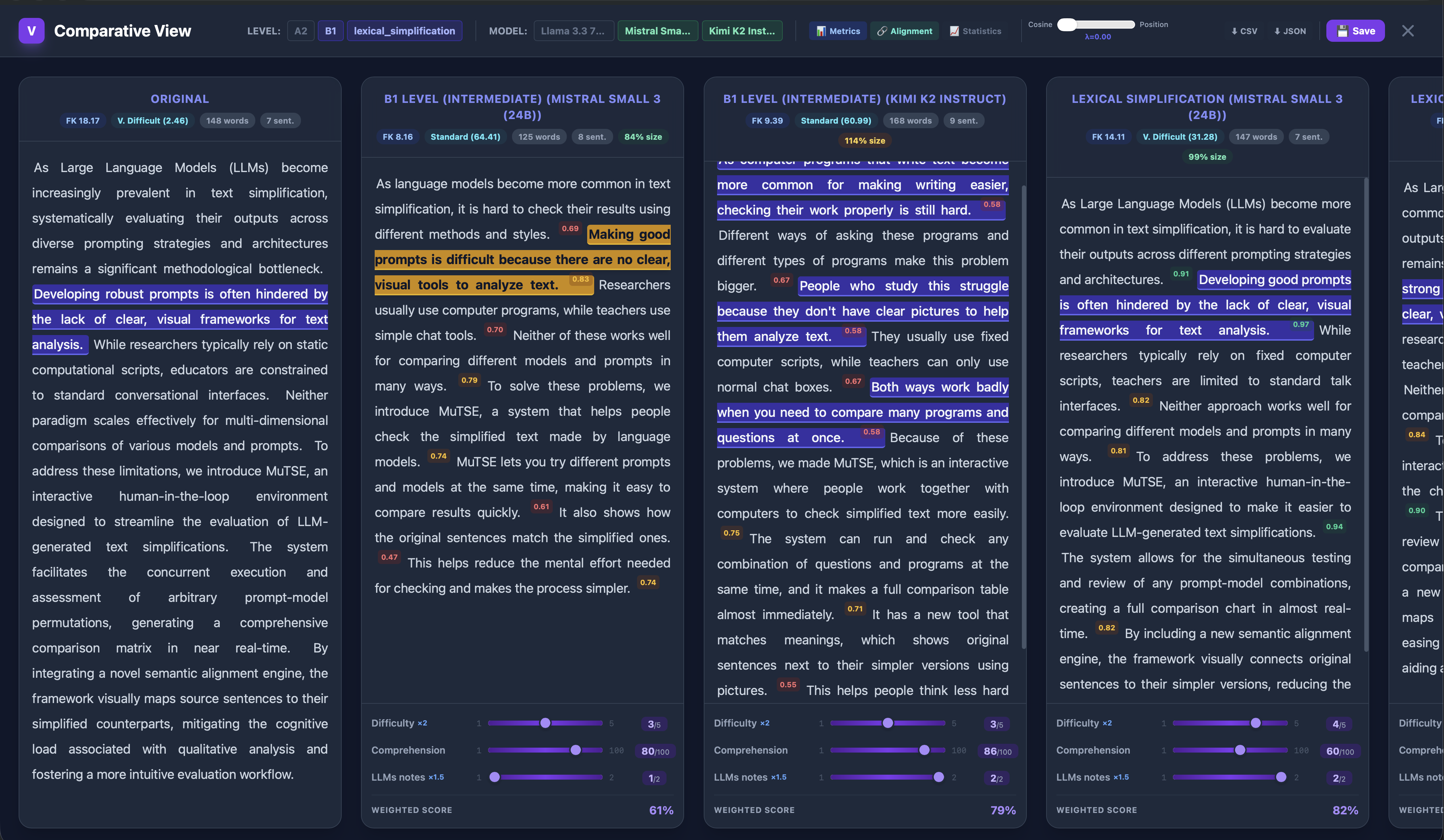}
    \\[2mm] % Spațiu mic între ele pentru a respira
    % A doua imagine (lambda = 2)
    \includegraphics[width=\textwidth]{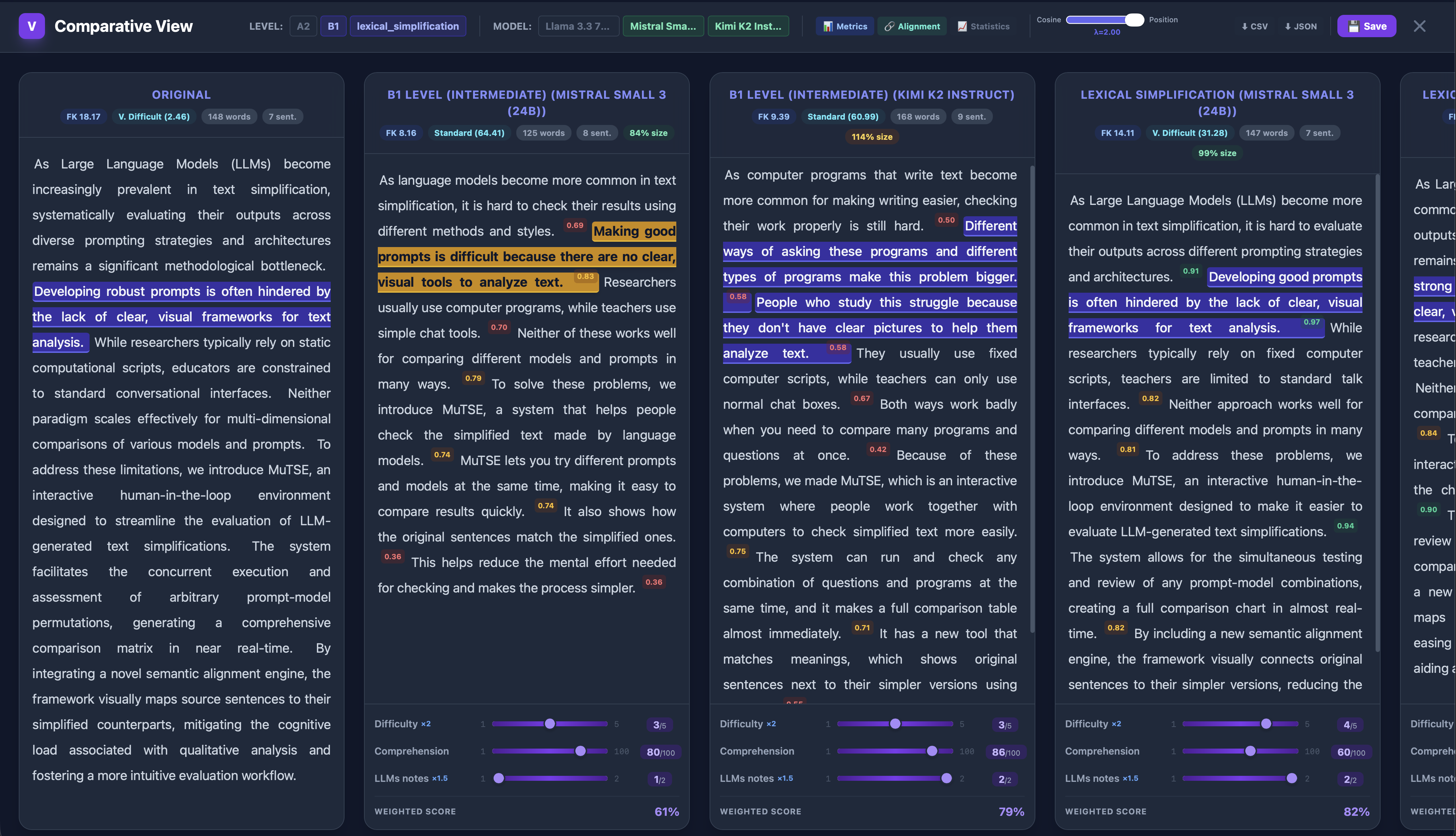}
    \caption{The impact of the Linearity Bias ($\lambda$) on the semantic alignment visualization. \textbf{Top:} Pure semantic matching ($\lambda = 0.00$) showing scattered, positionally distant alignments. \textbf{Bottom:} Strict positional penalty applied ($\lambda = 2.00$), resolving false positives and enforcing monotonic alignment.}
    \label{fig:lambda_comparison}
\end{figure}

\subsubsection{Computational Advantages of the Linearity Heuristic}
The integration of the linearity bias functions as a structural regularizer, effectively offsetting the reduced discriminative capacity inherent to heavily compressed embedding models. Whereas unpenalized semantic alignment ($\lambda = 0$) often mandates large-parameter models and dedicated GPU acceleration for optimal precision, our heuristic enables lower-dimensional embeddings to yield highly robust alignments using standard CPU architectures. This approach significantly lowers computational prerequisites, facilitating local deployment for educators and researchers without requiring specialized infrastructure. Moreover, by offloading the final alignment graph computation to the client interface using the pre-calculated similarity matrix, modifications to the $\lambda$ parameter are resolved instantaneously on the frontend without invoking redundant server-side processing.

\begin{figure}[htbp]
    \centering
    \begin{tikzpicture}[>=Latex, font=\sffamily\footnotesize, node distance=0.8cm and 1.5cm]
    
        \tikzset{
            base/.style={draw=black!70, rectangle, rounded corners, align=center, minimum height=0.8cm, thick, inner sep=4pt},
            input/.style={base, fill=blue!5},
            decision/.style={draw=black!70, diamond, aspect=1.8, align=center, fill=gray!5, thick, inner sep=1pt},
            matrix/.style={base, fill=blue!5, minimum height=1cm},
            process/.style={base, fill=gray!5},
            output/.style={base, fill=green!5}
        }
        
        % Nodes
        \node[input] (in) {Input: Original \&\\Simplified Sentences};
        \node[decision, right=of in] (dec1) {Tier 1: SBERT\\Valid vectors?};
        \node[decision, below=of dec1] (dec2) {Tier 2: TF-IDF\\Valid vectors?};
        \node[process, below=of dec2] (tier3) {Tier 3: Positional\\Zero Matrix};
        
        \node[matrix, right=of dec1] (basemat) {Base Cosine\\Matrix};
        \node[process, below=of basemat] (bias) {Apply Linearity\\Bias Penalty ($\lambda$)};
        \node[output, below=of bias] (out) {Final Alignment Graph\\(Many-to-one)};
        
        % Edges
        \draw[->, thick] (in) -- (dec1);
        \draw[->] (dec1) -- (dec2) node[midway, right] {No};
        \draw[->] (dec2) -- (tier3) node[midway, right] {No};
        
        \draw[->, thick] (dec1) -- (basemat) node[midway, above] {Yes};
        \draw[->, rounded corners] (dec2.east) -- ++(0.5,0) |- (basemat.west) node[pos=0.25, right] {Yes};
        \draw[->, rounded corners] (tier3.east) -- ++(1,0) |- (basemat.west);
        
        \draw[->, thick] (basemat) -- (bias);
        \draw[->, thick] (bias) -- (out);
        
    \end{tikzpicture}
    \caption{The 3-tier semantic alignment cascade showing hierarchical fallback logic and the application of Linearity Bias ($\lambda$).}
    \label{fig:alignment_engine}
\end{figure}
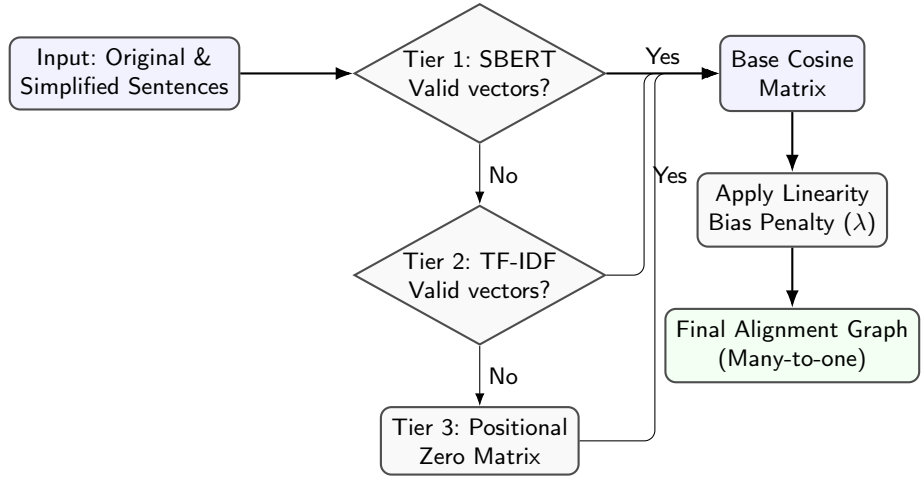

\section{Educational Features and Interactive Workflow}

The primary objective of MuTSE is to facilitate the qualitative evaluation of generated texts for educational purposes. To bridge the gap between algorithmic text generation and pedagogical utility, the system incorporates an interactive visualization interface, automated readability assessment models, and a customizable manual annotation framework.

\subsection{System Modularity: Prompts, Models, and Custom Criteria}

A core design principle of MuTSE is high modularity, accessible directly through the user interface's Settings module. This architecture permits researchers to dynamically configure the theoretical bounds of their evaluation matrix without altering the underlying application code:

\begin{itemize}
    \item \textbf{Prompt Management:} Users can define, persist, and retrieve custom instructional prompts. To facilitate immediate experimentation, the platform is pre-loaded with a curated set of 5 pre-defined prompts. While not necessarily state-of-the-art implementations, these foundational prompts serve as practical starting points for standard language proficiency objectives (e.g., A2, B1), encouraging standardized and reproducible research.
    \item \textbf{Model Integration:} The framework leverages cloud-based serverless infrastructure (via Together AI) to provide zero-provisioning access to diverse open-weight LLMs (e.g., Llama 3, DeepSeek V3, Qwen 2.5). Expanding the evaluation matrix is highly intuitive: users can dynamically add new models directly through the interface by simply inputting the model's official string identifier (e.g., \texttt{openai/gpt-oss-20b}) exactly as listed on the provider's public registry. This approach eliminates complex local hardware configurations while allowing users to instantly evaluate newly released models without altering the underlying codebase.
    \item \textbf{Customizable Annotation Framework:} To meaningfully capture the qualitative nuances of educational simplification, MuTSE features a highly configurable inline scoring system. Human evaluation in Natural Language Generation traditionally assesses dimensions such as \textit{Fluency} and \textit{Meaning Preservation}. However, recent literature highlights a pervasive lack of standardization and extreme diversity in evaluation methodologies and terminology across the field \cite{howcroft-etal-2020-twenty}. Furthermore, an ongoing methodological debate exists regarding rating scales: while discrete Likert scales (e.g., 1-5 points) are widely used, their application frequently deviates from best practices \cite{amidei-etal-2019-use}, prompting researchers to advocate for continuous measurement scales to capture finer quality distinctions \cite{graham-etal-2013-continuous}. Rather than enforcing a rigid evaluation paradigm, MuTSE operates as an evaluation-agnostic framework with zero predefined metrics. Through the Settings module, evaluators build their custom assessment criteria entirely from scratch. For each dimension, users establish their own custom numerical rating scale---supporting any arbitrary range from strict binary evaluations (e.g., a 1-to-2 scale representing True/False) up to continuous 100-point sliders---and assign a relative impact weight (from 0.1 to 10.0). During analysis, the system normalizes and aggregates these manual inputs to compute a weighted overall performance percentage for each generated column.
\end{itemize}

\subsection{Interactive Simplification and Visual Alignment}

The primary analytical interface of MuTSE is designed to mitigate the substantial cognitive load involved in evaluating multiple simplified derivations of a source text. Conventionally, tracking structural modifications, fragmentations, or semantic omissions across $M$ generated variants is a visually taxing process. 

MuTSE addresses this through a full-screen, side-by-side comparative layout, which directly visualizes the results of the multi-model generation process. To further manage visual complexity within high-dimensional comparisons, the interface features a dynamic filtering header. Evaluators can selectively toggle specific prompting strategies or LLM architectures on and off, instantly hiding or revealing their corresponding text columns. This functionality empowers users to compute a massive $P \times M$ evaluation matrix in a single execution, yet systematically analyze the outputs in smaller, focused subsets (e.g., isolating a single prompt across two specific models) according to their preferred analytical workflow without re-triggering generation tasks.

The interface then augments this focused view with an interactive cross-document highlighting mechanism governed by the pre-computed semantic alignment matrix (detailed in Section 3.2). Upon user interaction (hover or click) with a specific sentence in any document (source or generated), the system synchronously highlights the corresponding semantically aligned sentences across all active text columns. This alignment traversal operates natively within the client environment, ensuring near-instantaneous visual feedback and allowing evaluators to rapidly trace specific semantic units across disparate architectural outputs.

Crucially, the interface augments this qualitative visual tracking with rigorous, quantitative linguistic metrics and inline alignment diagnostics displayed directly alongside each variant column. Specifically, the system features a dedicated \textit{Alignment} visual toggle that embeds the exact cosine-similarity alignment score inline with each generated sentence. This granular visualization allows evaluators to immediately identify sentences with low correspondence to the original text, significantly accelerating the manual detection of unwanted LLM conversational artifacts (e.g., "Here is the simplified text:") or unprompted semantic hallucinations.

Furthermore, rather than relying exclusively on subjective visual inspection, researchers can continuously cross-reference human annotations with the system's calculated statistics. MuTSE computes and visually badges real-time textual diagnostics including:

\begin{itemize}
    \item \textbf{Textual Statistics:} Word frequency, total sentence count, and average sentence length.
    \item \textbf{Compression Ratio:} The proportion of retained lexical content relative to the source text, indicating structural expansion or reduction.
    \item \textbf{Flesch-Kincaid Grade Level:} An estimation of the US educational grade level required for comprehension \cite{Kincaid1975DerivationON}.
    \item \textbf{Flesch Reading Ease (FRE):} A quantitative assessment of overall reading accessibility on a 0-100 scale \cite{flesch1948new}.
\end{itemize}

By unifying these immediate, automated benchmarks—calculating metrics like syllable densities dynamically via language-specific hyphenation libraries—with the visual semantic mapping and inline custom scoring panels, MuTSE significantly reduces the evaluation time overhead while enhancing the consistency of multidimensional comparative analysis.

\begin{figure}[htbp]
    \centering
    \includegraphics[width=\textwidth]{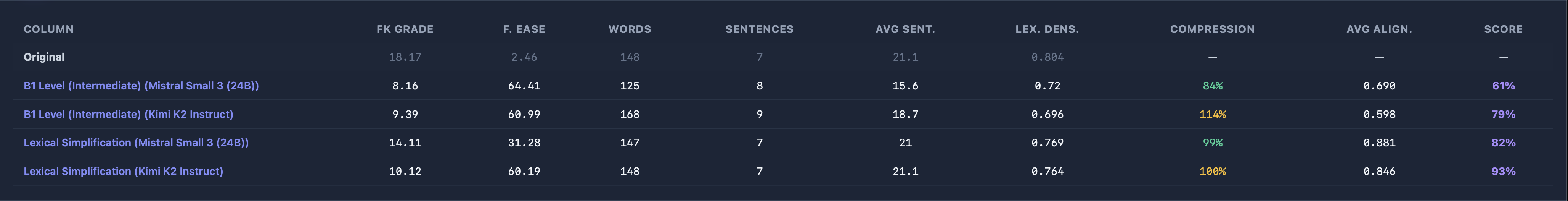}
    \caption{The integrated statistics module computing real-time educational metrics (e.g., Flesch-Kincaid, Reading Ease) and structural diagnostics across multiple prompt-model permutations.}
    \label{fig:statistics_table}
\end{figure}

\begin{figure}[htbp]
    \centering
    % Am redus lățimea la 85% din pagină pentru a preveni scalarea exagerată a înălțimii
    \resizebox{0.85\textwidth}{!}{%
    \begin{tikzpicture}[>=Latex, font=\sffamily\small, node distance=0.6cm and 0.5cm]
    
        \tikzset{
            base/.style={draw=black!70, rectangle, rounded corners, align=center, minimum height=0.8cm, minimum width=2.8cm, thick},
            input/.style={base, fill=blue!5},
            render/.style={base, fill=gray!15, minimum width=4cm},
            interaction/.style={base, fill=gray!5, dashed},
            reaction/.style={base, fill=gray!10},
            database/.style={draw=black!70, cylinder, shape border rotate=90, aspect=0.25, align=center, minimum height=1.2cm, minimum width=2.5cm, fill=green!5, thick}
        }
        
        % Nodes
        \node[input] (in) {Backend JSON Response};
        \node[render, below=0.6cm of in] (render) {Render Side-by-Side Comparison};
        
        % Path B (Center)
        \node[interaction, below=0.7cm of render] (pathB_act) {Path B: Adjust\\Linearity Bias ($\lambda$)};
        \node[reaction, below=0.5cm of pathB_act] (pathB_react1) {Client-side Graph\\Recomputation};
        \node[reaction, below=0.5cm of pathB_react1] (pathB_react2) {Update Visual\\Sentence Highlights};
        
        % Path A (Left)
        \node[interaction, left=0.8cm of pathB_act] (pathA_act) {Path A: Toggle\\Prompts/Models};
        \node[reaction, below=0.5cm of pathA_act] (pathA_react) {Instantly Hide/Reveal\\Columns};
        
        % Path C (Right)
        \node[interaction, right=0.8cm of pathB_act] (pathC_act) {Path C: Input\\Manual Scores};
        \node[reaction, below=0.5cm of pathC_act] (pathC_react) {Calculate Weighted\\Final Percentage};
        
        % Export
        \node[database, below=0.8cm of pathB_react2] (db) {Export\\(JSON/CSV Dataset)};
        
        % Edges
        \draw[->, thick] (in) -- (render);
        
        % Branching
        \draw[->, rounded corners] (render.south) -- ++(0,-0.4) -| (pathA_act.north);
        \draw[->] (render.south) -- (pathB_act.north);
        \draw[->, rounded corners] (render.south) -- ++(0,-0.4) -| (pathC_act.north);
        
        % Path A internal
        \draw[->] (pathA_act) -- (pathA_react);
        % Path B internal
        \draw[->] (pathB_act) -- (pathB_react1);
        \draw[->] (pathB_react1) -- (pathB_react2);
        % Path C internal
        \draw[->] (pathC_act) -- (pathC_react);
        
        % Converging to DB
        \draw[->, rounded corners] (pathA_react.south) |- (db.west);
        \draw[->] (pathB_react2.south) -- (db.north);
        \draw[->, rounded corners] (pathC_react.south) |- (db.east);
        
    \end{tikzpicture}%
    }
    \caption{The Vue.js frontend workflow demonstrating client-side interactivity, zero-latency recomputations, and parallel user interaction paths.}
    \label{fig:frontend_workflow}
\end{figure}
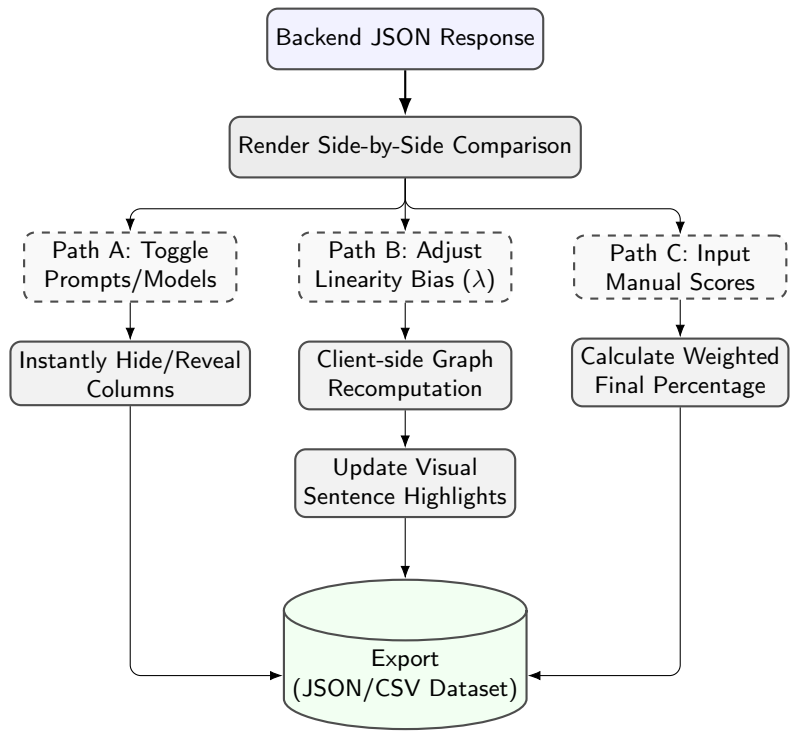

\subsection{Structured Data Export for NLP Research}

To support the creation of high-quality, human-in-the-loop datasets for future Natural Language Processing (NLP) research, all evaluation sessions are persistently stored locally. Crucially, researchers can export the entirety of an evaluation session---encompassing the source texts, generated outputs, semantic alignment links, readability metrics, and the weighted manual annotations---in standardized, structured formats (JSON and CSV).

This export functionality is essential for downstream applications. The structured CSV data is optimized for rapid Exploratory Data Analysis (EDA) and statistical significance testing, while the nested JSON format preserves the complex relational mapping of sentences and annotations, making it directly suitable for fine-tuning subsequent generations of language models or training specialized reward models.

\section{Limitations and Future Work}

While MuTSE significantly streamlines the evaluation of text simplification, the current architecture presents certain limitations. First, the reliance on a local JSON-based persistence layer, while maximizing portability for individual researchers, does not scale effectively for concurrent, multi-user deployments in large laboratory settings. Transitioning to a robust relational database is required for collaborative annotation campaigns. Second, while the integration of cloud-based LLM providers (e.g., Together AI) eliminates the need for local GPU clusters, users still face initial deployment friction regarding environment configuration (e.g., Python, Node.js dependencies).

For future work, the modularity of the semantic alignment engine presents significant opportunities beyond monolingual text simplification. The primary tier of our alignment cascade already utilizes a robust multilingual embedding model (\texttt{paraphrase-multilingual-MiniLM-L12-v2}) capable of processing over 50 languages, while the secondary TF-IDF tier captures cross-lingual morphological roots via character n-grams. Consequently, a natural extension of MuTSE is adapting the framework for Machine Translation (MT) evaluation and cross-lingual summarization. While the current implementation is highly optimized for simplification, verifying MT outputs would require recalibrating the linearity bias heuristic ($\lambda$), as cross-lingual syntactic restructuring may not strictly adhere to the monotonic sentence order typically observed in monolingual simplification. Expanding the interface to support these diverse Natural Language Generation evaluation branches remains a primary objective.

\section{Conclusion}

The systematic evaluation of Large Language Models in educational contexts requires tools that bridge the gap between automated generation and qualitative human assessment. In this paper, we introduced MuTSE, a human-in-the-loop web application designed to accelerate the comparative analysis of text simplification. By coupling an asynchronous multi-model generation pipeline with a novel, tiered semantic alignment engine, the system mitigates the cognitive load associated with tracking complex textual transformations. Furthermore, the integration of customizable annotation scales and automated readability metrics provides educators and applied linguists with an accessible, highly configurable environment. Ultimately, MuTSE not only streamlines the selection of optimal texts for language learners but also serves as a robust foundation for building high-quality annotated corpora for future NLP research.

\begin{credits}

\end{credits}
\bibliographystyle{splncs04}
\bibliography{references}





% LNCS Homepage, \url{http://www.springer.com/lncs}, last accessed 2023/10/25
% \end{thebibliography}
\end{document}